\documentclass[sn-mathphys-ay]{sn-jnl}% Math and Physical Sciences Numbered Reference Style 

\usepackage{hyperref}
\hypersetup{
	colorlinks   = true, %Colours links instead of ugly boxes
	urlcolor     = blue, %Colour for external hyperlinks
	linkcolor    = blue, %Colour of internal links
	citecolor    = blue %Colour of citations
}

\usepackage{natbib}
\usepackage{siunitx} % for aligning numbers by decimal points in table columns
\usepackage{graphicx}

\usepackage{blkarray}
\usepackage{amsmath}
\usepackage{mathtools}
\usepackage{bigstrut}
\usepackage{gauss}

\usepackage{stackengine}

\usepackage{adjustbox}
\usepackage{algorithm}
\usepackage{algpseudocode}

\usepackage{booktabs}
\usepackage{amssymb}
\usepackage{amsfonts}
\usepackage{amsmath}
\usepackage{graphicx}
\usepackage{rotating}
\usepackage{tabu}
\usepackage{float,lscape}
\usepackage{threeparttable}
\usepackage{dblfloatfix}
\usepackage{subcaption}
\usepackage{float}

\begin{document}

\title{Extending Explainable Ensemble Trees (E2Tree) to regression contexts}

\author[1]{\fnm{Massimo} \sur{Aria}}\email{massimo.aria@unina.it}

\author*[1]{\fnm{Agostino} \sur{Gnasso}}\email{agostino.gnasso@unina.it}

\author[1]{\fnm{Carmela} \sur{Iorio}}\email{carmela.iorio@unina.it}

\author[2]{\fnm{Marjolein} \sur{Fokkema}}\email{m.fokkema@fsw.leidenuniv.nl}

%\affil[2]{\orgdiv{Department}, \orgname{Organization}, \orgaddress{\street{Street}, \city{City}, \postcode{10587}, \state{State}, \country{Country}}}

\affil[1]{\orgdiv{Department of Economics and Statistics}, \orgname{University of Naples Federico II}, \state{Italy}}

\affil[2]{\orgdiv{Institute of Psychology}, \orgname{Leiden University}, \state{The Netherlands}}

%%==================================%%
%% Sample for unstructured abstract %%
%%==================================%%

\abstract{Ensemble methods such as random forests have transformed the landscape of supervised learning, offering highly accurate prediction through the aggregation of multiple weak learners. However, despite their effectiveness, these methods often lack transparency, impeding users' comprehension of how RF models arrive at their predictions. Explainable ensemble trees (E2Tree) is a novel methodology for explaining random forests, that provides a graphical representation of the relationship between response variables and predictors. A striking characteristic of E2Tree is that it not only accounts for the effects of predictor variables on the response but also accounts for associations between the predictor variables through the computation and use of dissimilarity measures. The E2Tree methodology was initially proposed for use in classification tasks. In this paper, we extend the methodology to encompass regression contexts. To demonstrate the explanatory power of the proposed algorithm, we illustrate its use on real-world datasets.}


%%\pacs[JEL Classification]{D8, H51}

%%\pacs[MSC Classification]{35A01, 65L10, 65L12, 65L20, 65L70}

\maketitle

\section{Introduction}
\label{sec:Introduction}
In today's data-centric world, Machine Learning (ML) methodologies play a pivotal role in decision-making processes across a broad spectrum of domains, such as finance and medicine. These advanced algorithms, while essential for parsing through big datasets and approximating complex patterns of association, often result in "black box" models. Such models, characterized by their inherent complexity, present significant challenges in terms of understandability and transparency. \\
Two main roads to improving understandability and transparency of ML can be distinguished (see e.g.  \citealp{rudin2019stop, molnar2020interpretable, conard2023spectrum}). On the one hand, \textit{interpretable} ML concerns itself with the intuition that an ML model's utility is intrinsically linked to the ease with which users can understand how the algorithm processes the input values to arrive at its prediction. Interpretable ML methods create models that are (assumed to be) directly and inherently interpretable by users, such as sparse regression, decision trees and rule ensembles \citep{strobl2009introduction, hastie2020best, fokkema2020fitting}. On the other hand, \textit{explainable} ML methods are used for post-hoc explanation of black-box models. Specifically, a black-box model is fitted and used for prediction, and predictions are then explained using post-hoc explanation tools, such as variable importances, partial dependence plots or Shapley values \citep{breiman2001random, hastie2009, lundberg2017unified}. Limitations of both approaches have been pointed out in the literature \citep{lipton2018mythos, rudin2019stop}, and it should be noted that the two terms are often used interchangeably. The distinction between interpretation and explanation, however, remains useful for distinguishing the different aims and approaches of a wide range of available tools. \\
%This segmentation underscores the spectrum of transparency rather than a binary distinction, emphasizing the nuanced nature of model comprehensibility and the challenges in achieving universal definitions or formalizations in explainability. 
%The main point of the issue lies in the interpretability and explainability of these machine learning systems. Despite the frequent interchangeable use of the two terms in scholarly works, they encapsulate distinct aspects of ML transparency as delineated by several seminal works (\citealp{gilpin2018explaining}; \citealp{lipton2018mythos}).
The imperative for explainable ML extends across all sectors where decision-making processes rely heavily on ML models, especially in scenarios marked by a disjunction between the model's explanations and the decision-maker's comprehension. E2Tree is a methodology designed for post-hoc explanation of black-box tree ensembles. It offers both local and global explanations, allowing users to understand the decision-making process of the model at both a local observation level and a global general level. This dual capability ensures that users can see the detailed logic for specific instances as well as the patterns and rules that the black-box model follows. E2Tree ensures that ML models remain tools of empowerment, offering an understandable decision-making process.\\
The current paper is organized as follows. Section~\ref{sec:Explainable machine learning} places E2Tree in the wider context of explainable ML. Our proposed extension of E2Tree to regression models is presented in Section~\ref{sec:proposal}. Experimental results are shown in Section~\ref{sec:empiricalresults}, and Section~\ref{sec:conclusions} concludes the paper.

\section{Explainable machine learning}
\label{sec:Explainable machine learning}
Explaining black-box models presents a range of challenges. First, these models, such as decision tree ensembles, are inherently opaque, shrouding their inner workings in complexity and making it difficult to provide transparent explanations for their predictions. Second, the limited interpretability of black-box models makes it difficult to understand how input features affect predictions, complicating efforts to interpret the underlying decision-making processes. In datasets with many features, elucidating the critical factors influencing a black-box model's decisions becomes daunting due to the multitude of interactions between features \citep{aria2021comparison}. This is exacerbated by the ability of black-box models to capture complex, non-linear relationships between features and the response.  Further, there is an inherent risk of inheriting biases from training data, making it critical to explain and correct for such biases despite the complexity of the models \citep{bouazizi2024enhancing}. In addition, the computational overhead of generating explanations for black-box models places practical limits on the scalability and applicability of explanation methods. Finally, the lack of standardized methods and metrics for evaluating explanation quality promotes ambiguity and hinders the comparison and evaluation of different explanation approaches and their effectiveness \citep{singh2021explainable}.\\
%Explainable machine learning (XAI) is an interdisciplinary field of artificial intelligence (AI) that focuses on providing insights into the decision-making process of ML models \citep{spyrison2024exploring}.
Explainable ML methods are used to explain the inner logic of ML decision-making processes. They bring to light the rationale behind a model's predictions by providing an understanding of the decision-making process of different ML systems. When the strengths and weaknesses of an ML system to make a decision or give some recommendations are well known, the performance of the existing system can be improved \citep{vilone2021classification}. Explainable machine learning also helps build trustworthy and transparent systems for a wide range of applications, especially critical applications such as medical decision-making. Explainable machine learning methods seek to make ML models more transparent and understandable to humans, allowing users to gain a better understanding of how and why these models make the predictions they do. By providing explanations for predictions of black-box methods, explainable ML aims to increase trust and reliance on these models while also promoting responsible ML practices \citep{guidotti2021evaluating}. \\
With the aim to obtain explainability from tree ensemble models, a new method called E2Tree has been recently introduced by \cite{aria2024explainable}. Initially proposed for classification tasks, E2Tree offers a coherent, tree-like structure that facilitates an in-depth understanding of the relationships between features and their corresponding predictions while preserving the model's predictive accuracy. The proposed method provides a global explanation applicable to the entire dataset, highlights pivotal features, and effectively visualizes interactions between them.
%By merging the interpretability of individual decision trees with the stability provided by ensemble models, \textit{E2Tree} delivers precise and accessible explanations.
By integrating the interpretability of individual decision trees with the stability afforded by ensemble models, E2Tree offers comprehensive and readily accessible explanations.

\section{E2Tree in the regression context}
\label{sec:proposal}

Building on the foundations recently laid in in \cite{aria2024explainable}, we extend the explanatory power of E2Tree to the regression setting. E2Tree demonstrated high capabilities in the explanation of ensemble tree structures in classification contexts. The current work aims to provide comparable benefits for continuous outcome models.
In accordance with the methodology proposed in \cite{aria2024explainable}, the core aim is to obtain an explainable representation of an RF model in the form of a single tree-like structure.\\
\noindent Let $H$ be a random forest (RF) model composed of $B$ weak learners
$$
H = \{ H_{1},H_{2},\ldots,H_{b},\ldots,H_{B} \}.
$$
Let $\{(X_{1},Y_{1}),\ldots,(X_{n},Y_{n}) \}$ be a sample of size $n$ used to train the RF. Let $u_{i} = \{y_{i},x_{i}\}$ and $u_{j} = \{y_{j},x_{j}\}$ be the vectors of observed values of the $i$-th and 
$j$-th observations, respectively.\\
E2Tree starts from the definition of a dissimilarity matrix:\\
%(Matrix of Response Means or matrix of target Averages??):\\

\begin{figure}[H]
\[
D = \qquad
\begin{gmatrix}[b]
	\mathllap{u_1\quad}~~~0      &  d_{12} & d_{13} & \cdots & d_{1i} & d_{1j} & \cdots & d_{1n} \\
	\mathllap{u_2\quad} d_{21} &   0     & d_{23} & \cdots & d_{2i} & d_{2j} & \cdots & d_{2n} \\
	\mathllap{u_3\quad} d_{31} & d_{32}  & 0   & \cdots & d_{3i} & d_{3j} & \cdots &    d_{3n} \\
	\vdots & \vdots & \vdots & \ddots & \vdots & \vdots & \ddots & \vdots \\
	\mathllap{u_i\quad}d_{i1}&d_{i2} & d_{i3} &\cdots & 0 & d_{ij} & \cdots & d_{in} \\
        \mathllap{u_j\quad}d_{j1}&d_{j2} & d_{j3} & \cdots & d_{ij} & 0 & \cdots & d_{jn} \\
	\vdots & \vdots & \vdots & \ddots & \vdots & \vdots & \ddots & \vdots \\
	\mathllap{u_n\quad}d_{n1}&d_{n2} & d_{n3} & \cdots & \cdots & d_{nj} & \cdots & 0 \colops
		\def\colmultlabel#1{\makebox[1.2em]{$#1$}}
	\mult0{u_1}
	\mult1{u_2}
	\mult2{u_3}
	\mult4{u_i}
        \mult5{u_j}
	\mult7{u_n}
\end{gmatrix}
 = 1-
\begin{bmatrix}
 & &  &  &  &  & & \\
 & &  &  &  &  & &  \\
 & &  &  &  &  & & \\
 & &  &  O_{ij}  &  & & \\
 & &  &  &  &  & & \\
 & &  &  &  &  & & \\
 & &  &  &  &  & &
\end{bmatrix}
\]
\end{figure}
\noindent The matrix $D$ can be interpreted as a fuzzy partition of the data obtained by the RF procedure. In this context, $d_{ij}$ measures the membership value of observations $i$ and $j$ belonging to the same group. This results in a novel representation of the RF model, where $O_{ij}$ is a weighted co-occurrence measure between observations $u_{i}$ and $u_{j}$ along the sequence $H_{1},\ldots,H_{B}$. Hence, the values of matrix $O$ measure how often a pair of observations fall into the same node in an RF.
%where $O_{ij}$ is a weighted co-occurrence measure between observations $u_{i}$ and $u_{j}$ along the sequence $H_{1},\ldots,H_{B}$. \\
%In the matrix $D$, $d_{ij}$ can be interpretated as the membership probabilities of $i$-th and $j$-th observations to belong to the same tree resulting from the RF procedure. Thus, the matrix $D$ partition $u_{i}$ and $u_{j}$ in $H_b$ fuzzy overlapping clusters.
%This resulting in a novel representation of the RF model, where the values of $O$ measure how often a pairs of observations fall into the same node in an RF.
We define $O_{ij}$ as:
\begin{equation}
    O_{ij} = \frac{\sum_{b=1}^{B}I(u_{i} \wedge u_{j}) \cdot W_{(t)_{ij}|b}}{\max(\sum_{b}u_{i}W_{(t)_{i}|b};\sum_{b}u_{j}W_{(t)_{j}|b})},
\quad \quad 0 \leq O_{ij} \leq 1
\label{eq1}
\end{equation}

\noindent where $W_{(t)_{ij}|b}$ is a local goodness of fit measure of node $t$ of the $b$-th learner. \\
$W_{(t)_{ij}|b}$ is a normalized generic measure of local goodness of fit at node $t$ such that $0 \leq W_{(t)_{ij}|b} \leq 1$.\\
In the regression context, we propose the following local goodness of fit measure:

\begin{equation*}
   W_{(t)_{ij}|b} = 
    \begin{cases}
  0 & \text{if } 1-\mathit{NMSE}_{(t)_{ij}|b} \leq 0 \\
  1-\mathit{NMSE}_{(t)_{ij}|b} & \text{otherwise}
\end{cases}
\label{eq2}
\end{equation*}

\noindent where $\mathit{NMSE}_{(t)_{ij}|b}$ is the normalized measure of the mean square error at node $t$:

\begin{equation}
\mathit{NMSE}_{(t)_{ij}|b} = \frac{\mathit{MSE}_{(t)_{ij}|b}}{\mathit{MSE}_{max}\cdot p(t)},
\label{eq3}
\end{equation}

\noindent where $\mathit{MSE}_{(t)_{ij}|b}$ is the mean square error at node $t$, $p(t)=n(t)/N$ is the proportion of observations falling into the $t$-th node,  and $\mathit{MSE}_{max}$ is the theoretical maximum of the $\mathit{MSE}$ obtained under the assumption formulated by \cite{jacobson1969}:
\begin{equation*}
    \mathit{MSE}_{max} = \frac{(\max(Y|b)-\min(Y|b))^2}{9}.
\end{equation*}

In the literature, several upper bounds on variance as a measure of maximum dispersion from the mean have been proposed that also require knowledge about the underlying distributions. Here, variance bounds themselves are not the principal subjects of interest, but we refer the reader to \cite{samuelson1968deviant}, \cite{arnold1974schwarz} and \cite{seaman1985maximum}.\\
Alternative measures for $W_{(t)_{ij}|b}$ can be based on different approaches to normalize $\mathit{NMSE}$ at node $t$. For example, following the concept of coefficient of variation, an alternative proposal for $\mathit{NMSE}$ could be: 
\begin{equation*}
\mathit{NMSE}_{(t)_{ij}|b} = \frac{\mathit{\mathit{MSE}}_{(t)_{ij}|b}}{\mid{\hat{Y}_t}\mid}.
\end{equation*}
In alignment with the methodology proposed by \cite{aria2024explainable}, we also establish a set of stopping criteria that reflect the traditional approaches of tree estimation. In particular, we consider thresholds related to the size and impurity of terminal nodes. It is well-known that RFs tend to include many non-informative splits due to the random selection of split variables. Consequently, it can be expected that with an increasing number of nodes in a tree, the dissimilarity values $d_{ij}$ between two observations $(i, j)$ concerning a given classifier will tend to converge to one. To address this issue, our proposed algorithm incorporates two additional stopping rules. After identifying the optimal split based on dissimilarity, we verify that the value of $\mathit{NMSE}_{{(t)_{ij}|b}}$ meets a specified threshold $\gamma$. The node $t$ becomes terminal if $\mathit{NMSE}_{{(t)_{ij}|b}} \leq \gamma$. Furthermore, for each split, the distributions between the child nodes are verified to be identical through the Mann-Whitney test, as proposed by \cite{mann1947test}. If the predicted values demonstrate identical distributions in the child nodes after the splitting, the branch is removed and the parent node is regarded as a terminal node. Otherwise, the splitting procedure proceeds.\\

\begin{algorithm}[H]
	\renewcommand{\thealgorithm}{}
	\caption{E2Tree}
	\label{e2tree}
	\begin{algorithmic}[1]
		\Procedure{}{}     
		\State System Initialization:
		\State \quad - Calculate the dissimilarity matrix $D$ from the RF model
		\State \quad - Set Stopping Rules		
		\State \quad - Generate Matrix $S$ of all possible splits
		\Repeat
		\State \quad - Select node $t$
		\State \quad - Identify best split $s^*$ 
		\State \quad - Generate the two children nodes 
		\State \quad \quad $t_L=t\cdot2$
		\State \quad \quad $t_R=(t\cdot2)+1$
		\State \quad - Calculate statistics for $t_L$ and $t_R$ 
		\Until{At least one of the stopping rules is satisfied} 
		
		\EndProcedure
	\end{algorithmic}
\end{algorithm}

%%%%%%%%%%%%%%%%%%%%%%%%%%%%%%%%%%%%%%%%%%%%%%%%%%
\section{Illustrative examples}
\label{sec:empiricalresults}
To evaluate the performance of our proposal, we performed applications on freely available data sets from UCI Machine Learning repository. Firstly, we employed a simplified illustrative example with the Iris dataset \citep{iris} to elucidate how E2Tree works. Then, we analyze the Auto MPG data set \citep{auto}. We discuss the impressive performances achieved by E2Tree, which also explains the decision-making process of a RF for regression.\\
Data analysis was performed using the \texttt{R} software with our \textit{e2tree} package, which is available at the \texttt{GitHub} repository \url{https://github.com/massimoaria/e2Tree}.\\
We build up the RF model by partitioning the datasets into a training set and a testing set. Following the methodology outlined by \cite{guyon1997}, $70\%$ of each dataset was allocated to the training set, while the remaining $30\%$ was reserved for testing. The RF model was configured with 500 trees, and at each node  $p/3$  randomly selected features were evaluated. In our algorithm, we set the specified threshold $\gamma=0.05$.

\subsection{A toy example: Iris data}
The Iris dataset is a well-known dataset in the machine learning community. It comprises 150 samples from three species of Iris flowers: Iris setosa, Iris versicolor, and Iris virginica. Each species has 50 observations, making the dataset balanced with respect to the qualitative feature Species. Each sample has four quantitative features, with summary statistics presented in Table~\ref{irisattr}.

\begin{table}[h!]
\centering
\begin{tabular}{lS[table-format=1.2]S[table-format=1.2]S[table-format=1.1]S[table-format=1.1]}
  \hline
Feature & {Mean} & {SD} & {Min} & {Max} \\ 
  \hline
Sepal's Width & 3.06 & 0.44 & 2.0 & 4.4 \\ 
Sepal's Length & 5.84 & 0.83 & 4.3 & 7.9 \\ 
Petal's Width & 1.20 & 0.76 & 0.1 & 2.5 \\ 
Petal's Length & 3.76 & 1.77 & 1.0 & 6.9 \\ 
   \hline
\end{tabular}
\caption{Summary statistics of the Iris dataset attributes.}
\label{irisattr}
\end{table}

Here, we focus on running a regression on the variable Petal Length.\\
We built the RF regression model using the training set, as previously described. Table~\ref{tab:rf_results_iris} summarizes the RF output.

\begin{table}[h!]
\centering
\begin{tabular}{ll}
\hline
Type of random forest:   & regression \\ 
Number of trees:          & 500        \\ 
No. of variables tried at each split: & 1 \\ 
Mean of squared residuals: & 0.084 \\ 
\% Var explained: & 97.49 \\ 
\hline
\end{tabular}
\caption{Iris data: Results of random forest regression.}
\label{tab:rf_results_iris}
\end{table}

%\vspace{-1cm}

Table~\ref{tab:feature_importance_iris} shows the feature importance of the RF model output, as measured by Percentage Increase in Mean Squared Error (\%IncMSE) and Increase in Node Purity (IncNodePurity). The former is a reliable and informative measure, with higher values indicating greater feature importance. The latter, related to the loss function (MSE for regression), quantifies how much splits on each predictor variable contribute most to the overall decrease in intra-node variance. 

\begin{figure}[H]
	\centering
	\includegraphics[scale=0.25]{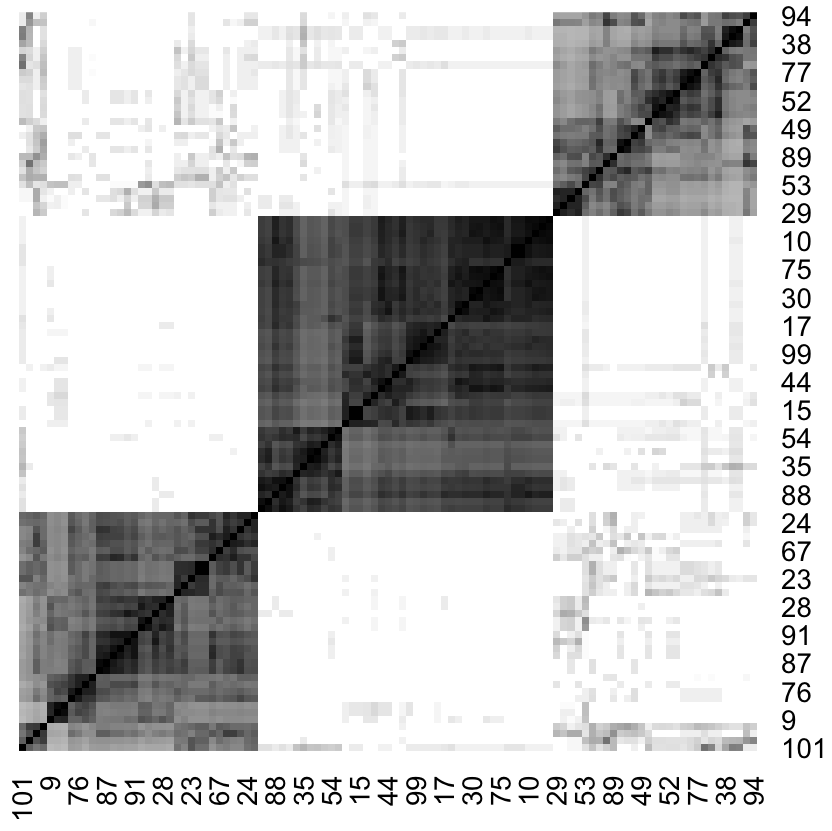}
	\caption{Iris data: heatmap of the matrix $O_{ij}$.}
	\label{fig:heatmap_iris}
\end{figure}

\begin{table}[h!]
    \centering
    \begin{tabular}{lS[table-format=3.2]S[table-format=3.2]}
        \hline
        Feature & {\%IncMSE} & {IncNodePurity} \\
        \hline
        Petal Width  & 2.25 & 108.03 \\
        Species      & 2.18 & 108.04 \\
        Sepal Length & 0.92 & 86.34 \\
        Sepal Width  & 0.29 & 36.40 \\
        \hline
    \end{tabular}
\caption{Iris data: Feature importance metrics in RF.}
\label{tab:feature_importance_iris}
\end{table}
The initial stage in the generation of the E2tree is the construction of  $O_{ij}$, as defined in Equation \ref{eq1}, using the outcomes produced by the RF algorithm. The heatmap of the matrix $O_{ij}$ is illustrated in Figure~\ref{fig:heatmap_iris}.  The matrix $O_{ij}$ enables us to quantify the frequency with which pairs of observations co-occur within the same node of an RF, with the result weighted by the normalised generic measure of local goodness of fit, $W_{(t)_{ij}|b}$. The normalised mean square error, $\mathit{NMSE}_{(t)_{ij}|b}$, is computed in accordance with the procedure set out in Equation \ref{eq3}. Given that E2Tree is based on the concept of a dissimilarity matrix $D$, this matrix is calculated next as $1 - 
O_{ij}$ (Section~\ref{sec:proposal}).\\
%weighted by the normalized generic measure of local goodness of fit $W_{(t)_{ij}|b}$, where $\mathit{NMSE}_{(t)_{ij}|b}$ is computed as in Equation \ref{eq3}. 
%
%
%Since E2Tree starts from the definition of a dissimilarity matrix $D$, the next step is the computation of $D$, as defined in the Section~\ref{sec:proposal}. 

\begin{figure}[H]
	\centering
	\includegraphics[scale=0.25]{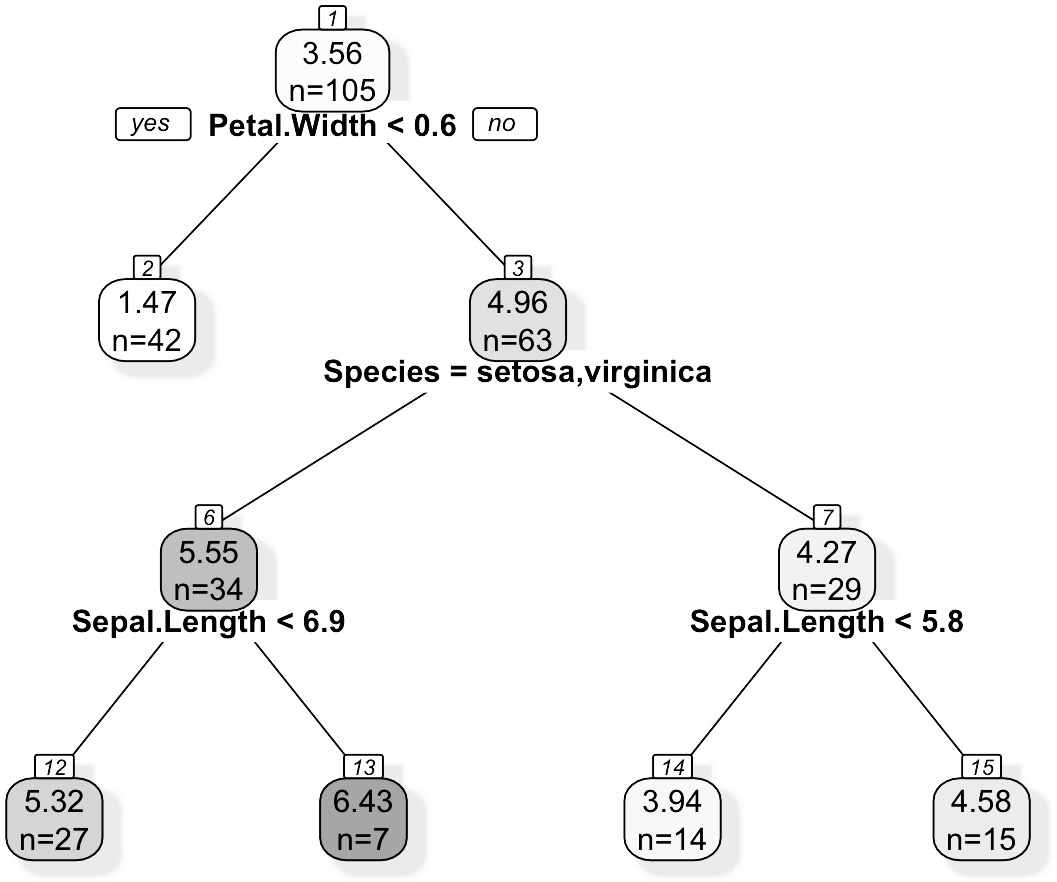}
	\caption{Iris data: Path visualization of E2Tree. The color intensity indicates the magnitude of the predicted values: darker shades typically correspond to higher predicted values, while lighter shades correspond to lower predicted values.}
	\label{fig:rplotiris}
\end{figure}

%E2Tree is shown in Figure~\ref{fig:rplotiris} and jointly with Table~\ref{tab_res_iris} provides insight into the path leading to each terminal node in the E2Tree, explaining the predictions made by the RF. E2Tree not only accounts for the effects of predictor variables on the response but also incorporates the associations between predictor variables through the computation and use of dissimilarity measures.
%
Figure~\ref{fig:rplotiris} and Table~\ref{tab_res_iris} illustrate the E2Tree and provide insight into the path leading to each terminal node, thereby elucidating the predictions made by the RF. Note that E2Tree naturally handles both numerical or ordinal predictors (Petal.Width and Sepal.Length, in this example), as well as ordered-categorical predictors (Species, in this example). In addition to accounting for the effects of predictor variables on the response, E2Tree also incorporates the associations between predictor variables through the computation and utilisation of dissimilarity measures. 

\begin{sidewaystable}
        \centering
        \begin{adjustbox}{max width=\textwidth}
    \begin{tabular}{|l|l|l|l|l|l|l|}
    \hline
        $t$ & $n_{t}$ & Prediction & $W_{t_{ij}}$ & $s^*$ & Node Type & Path \\ \hline
        1 & 105 & 3.56 & 0.11 & Petal.Width $\leq$ 0.6 & Non-Terminal & ~ \\ \hline
       2 & 42 & 1.47 & 0.98 & - & Terminal & Petal.Width $\leq$ 0.6 \\ \hline
        3 & 63 & 4.96 & 0.70 & Species = (setosa, virginica) & Non-Terminal & Petal.Width $>$ 0.6 \\ \hline
        6 & 34 & 5.55 & 0.73 & Sepal.Length $\leq$ 6.9 & Non-Terminal & Petal.Width $>$ 0.6 \& Species = (setosa, virginica) \\ \hline
        12 & 27 & 5.32 & 0.87 & - & Terminal & Petal.Width $>$ 0.6 \& Species = (setosa, virginica) \& Sepal.Length $\leq$ 6.9 \\ \hline
        13 & 7 & 6.43 & 0.60 & - & Terminal & Petal.Width $>$ 0.6 \& Species = (setosa, virginica) \& Sepal.Length $>$ 6.9 \\ \hline
        7 & 29 & 4.27 & 0.82 & Sepal.Length $\leq$ 5.8 & Non-Terminal & Petal.Width $>$ 0.6 \& Species = versicolor  \\ \hline
        14 & 14 & 3.94 & 0.81 & - & Terminal & Petal.Width $>$ 0.6 \& Species = versicolor \& Sepal.Length $\leq$ 5.8 \\ \hline
        15 & 15 & 4.58 & 0.87 & - & Terminal & Petal.Width $>$ 0.6 \& Species = versicolor \& Sepal.Length $>$ 5.8 \\ \hline
   \end{tabular}
    \end{adjustbox}
    \caption{Iris data: results of E2Tree.}
    \label{tab_res_iris}
\end{sidewaystable}

\begin{figure}[H]
\centering
\begin{subfigure}{0.45\textwidth}
  \centering
  \includegraphics[scale=0.18]{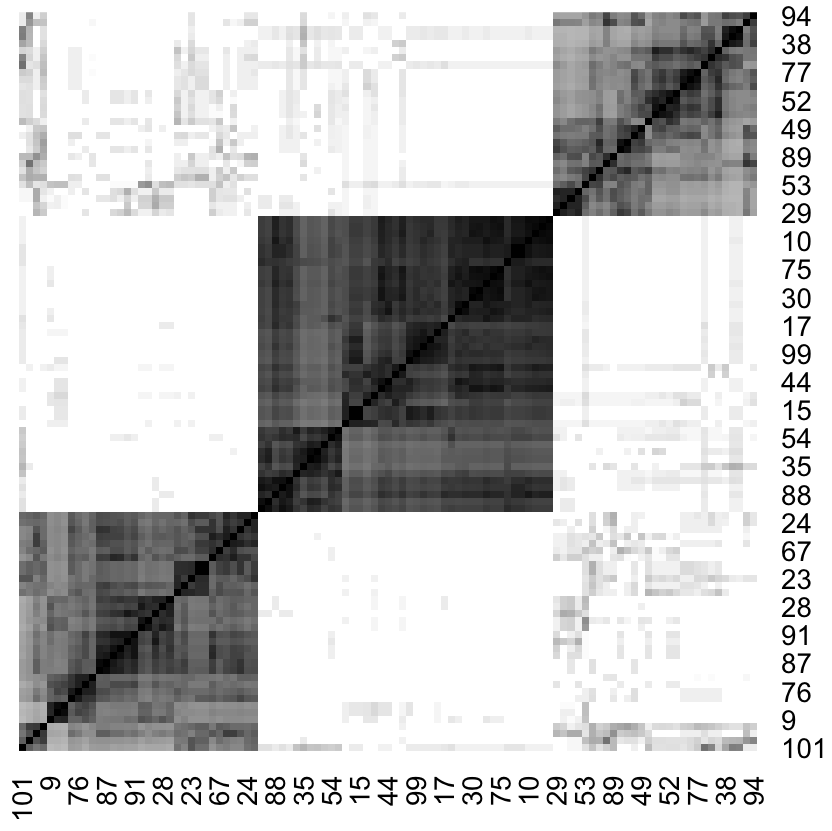}
  \caption{}
\end{subfigure}
\hfill
\begin{subfigure}{0.45\textwidth}
  \centering
  \includegraphics[scale=0.18]{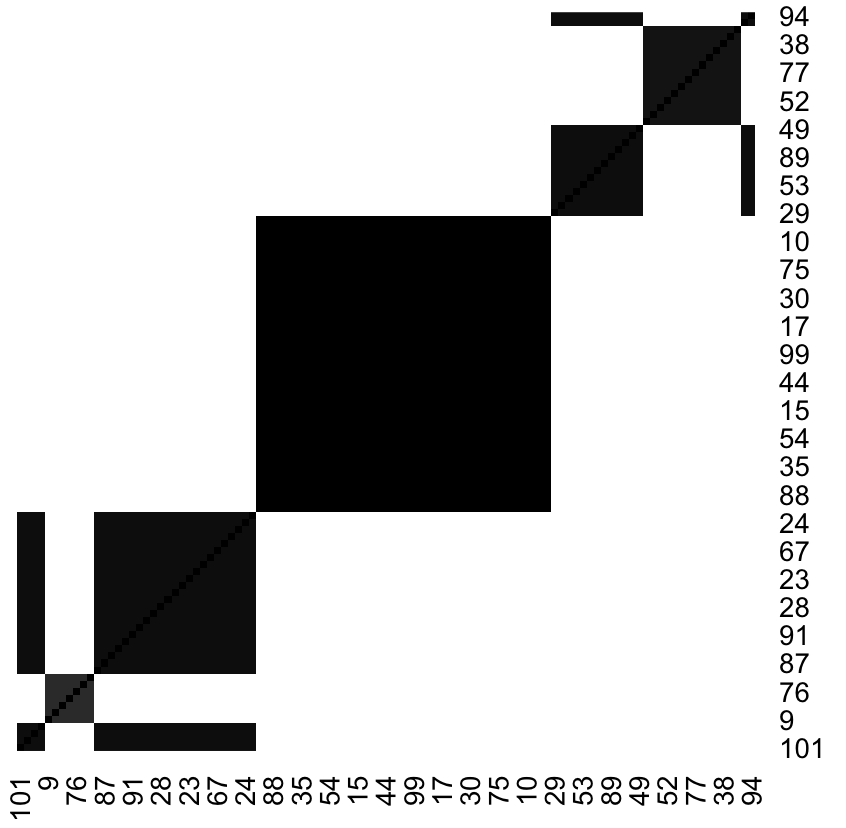}
  \caption{}
\end{subfigure}
\caption{Iris data: Heatmap of the matrix $O_{ij}$ (sub-plot a) and heatmap of the matrix $\hat{O}_{ij}$ estimated by E2Tree (sub-plot b).}
\label{comparison1}
\end{figure}

%To demonstrate E2Tree’s effectiveness in explaining the internal workings of RF, Figure~\ref{comparison1} presents a comparison between the heatmap of the matrix $O_{ij}$ obtained from the RF output and heatmap of the matrix $\hat{O}_{ij}$ estimated by E2Tree.
In order to demonstrate the efficacy of E2Tree in explaining the internal workings of RF, Figure~\ref{comparison1} presents a comparison between the heatmap of the matrix $O_{ij}$ obtained from the RF output and the heatmap of the matrix $\hat{O}_{ij}$ estimated by E2Tree.

\begin{figure}[H]
	\centering
	\includegraphics[scale=0.6]{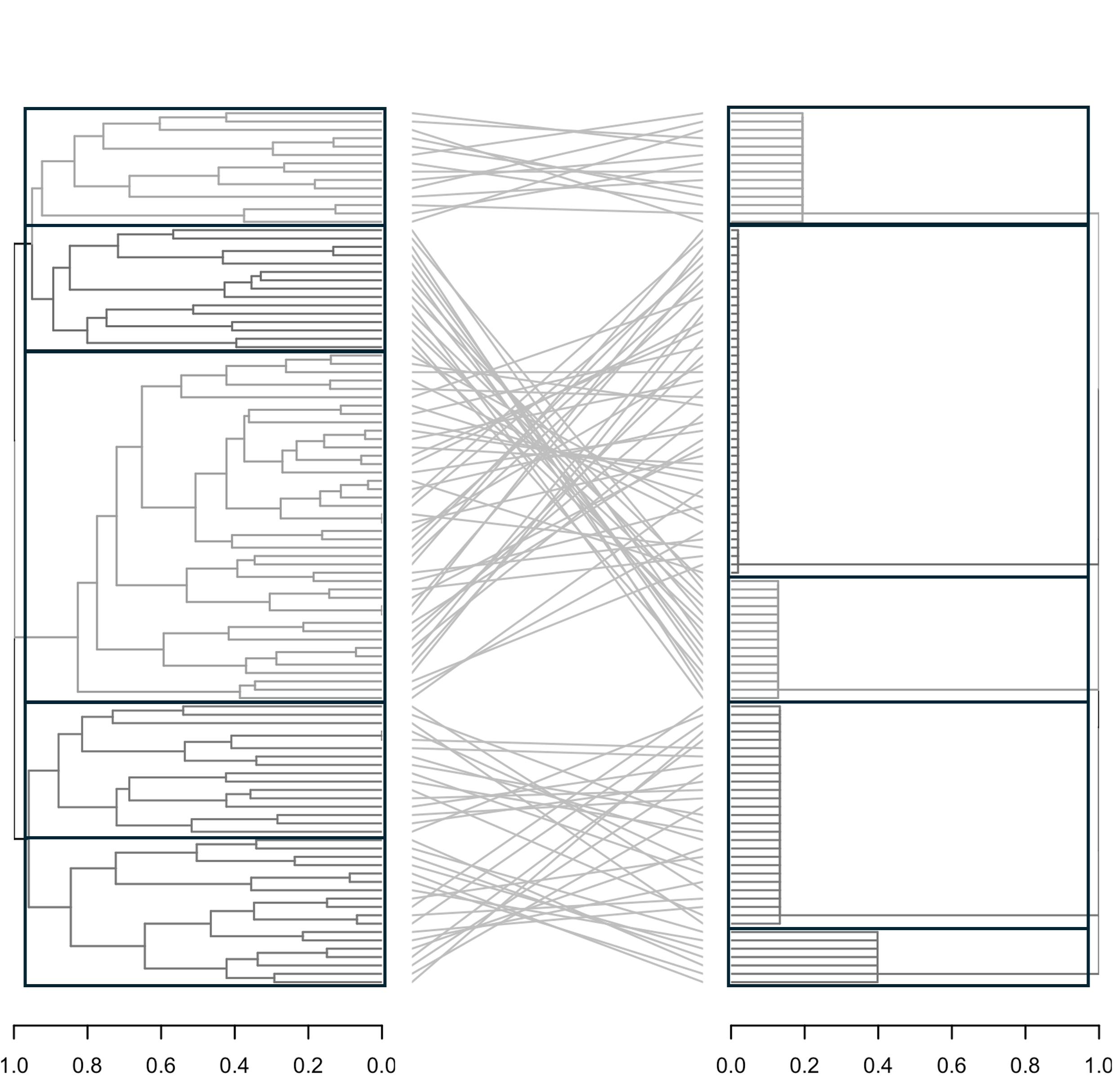}
	\caption{Iris data: Comparison between partitions ($k=5$) obtained applying hierarchical clustering analysis on $O_{ij}$ (left-hand side) and $\hat{O}_{ij}$ (right-hand side).} 
	\label{fig:rplotiriclus}
\end{figure}

To evaluate whether the matrix $\hat{O}_{ij}$ estimated by E2Tree is able to reconstruct the structure of the matrix $O_{ij}$ obtained from the results of the RF model, we performed a hierarchical cluster analysis on $O_{ij}$ and $\hat{O}_{ij}$ with complete linkage criterion.
The number of clusters we used is equal to $k=5$, where $k$ are the terminal nodes in E2Tree (see Figure~\ref{fig:rplotiriclus}). \\
To verify if $\hat{O}_{ij}$ accurately reflects the structure of $O_{ij}$, we computed the Fowlkes–Mallows index (FMI) \citep{fowlkes1983method}. The FMI is a widely used metric for comparing the similarity of two clustering results. It ranges from 0 to 1, where a value of 1 indicates perfect agreement between the two clusterings, and a value of 0 indicates no agreement beyond what would be expected by chance. 
The FMI was found to be approximatively equal to 90\%, indicating a high degree of similarity between $O_{ij}$ and $\hat{O}_{ij}$. This result suggests that E2Tree effectively captures the structure of the RF model.
%effectively captures the structure present in the RF model's outcomes.

%%%%%%%%%%%%%%%%%%%%%%%%%%%%%%%%%%%%%%%%%%%%%%%%%%%%%%%%%%%%%%%%%%%%%%%

\subsection{Auto MPG data set}

The Auto MPG dataset contains 398 entries, each representing a unique automobile model. The dataset is designed to predict the fuel efficiency of cars, measured in miles per gallon (MPG), using several predictor variables. These variables include quantitative features (see Table~\ref{tab:auto_mpg_stats}). The dataset provides a set of attributes, making it suitable for regression analysis and other machine learning tasks.

\begin{table}[h!]
\centering
\begin{tabular}{lS[table-format=4.4]S[table-format=4.4]S[table-format=4.4]S[table-format=4.4]}
  \hline
Feature & {Mean} & {SD} & {Min} & {Max} \\ 
  \hline
MPG & 23.51 & 7.82 & 9.0 & 46.6 \\ 
Cylinders & 5.45 & 1.71 & 3 & 8 \\ 
Displacement & 193.41 & 104.27 & 68 & 455 \\ 
Horsepower & 51.39 & 38.49 & 46 & 230 \\ 
Weight & 2970 & 849.40 & 1613 & 5140 \\ 
Acceleration & 15.57 & 2.76 & 8.0 & 24.8 \\ 
Model Year & 76.01 & 3.70 & 70 & 82 \\ 
Origin & 1.57 & 0.81 & 1 & 3 \\ 
   \hline
\end{tabular}
\caption{Summary statistics of the Auto MPG dataset attributes.}
\label{tab:auto_mpg_stats}
\end{table}

Here, we conduct a regression analysis on the MPG variable. We performed the RF regression model utilizing the training set as before. The  RF results are summarized in Table~\ref{tab:rf_results_cars}.

\begin{table}[h!]
\centering
\begin{tabular}{ll}
\hline
Type of random forest:   & regression \\ 
Number of trees:          & 500        \\ 
No. of variables tried at each split: & 2 \\ 
Mean of squared residuals: & 8.10 \\ 
\% Var explained: & 85.94 \\ 
\hline
\end{tabular}
\caption{Auto MPG data: Results of random forest regression.}
\label{tab:rf_results_cars}
\end{table}

%Figure~\ref{fig:rplotcars} and Table~\ref{tab_res_cars} present the output of our proposed E2Tree model on the Auto MPG dataset.

\begin{figure}[H]
	\centering
	\includegraphics[scale=0.15]{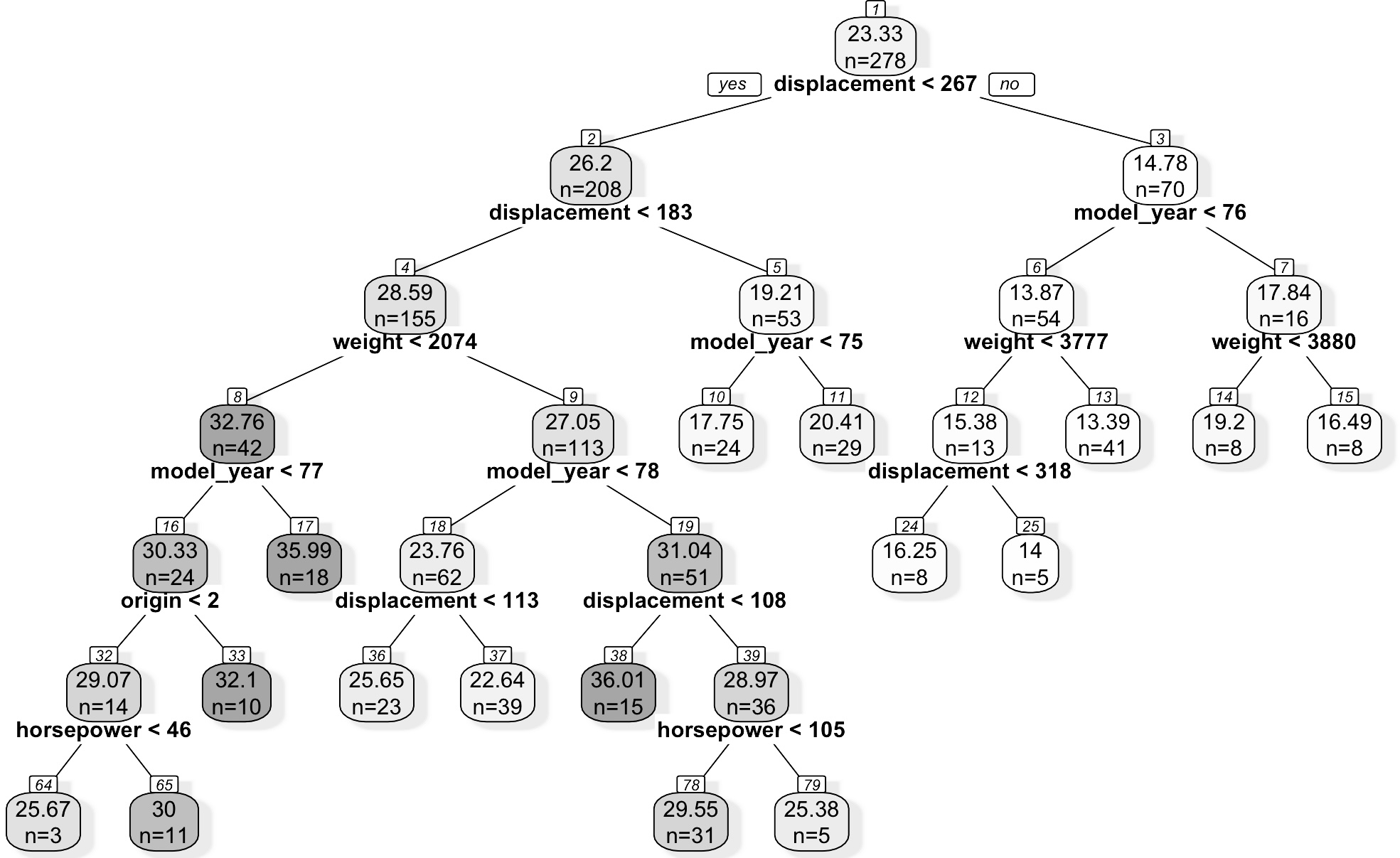}
	\caption{Auto MPG data: path visualization of E2Tree. The color intensity indicates the magnitude of the predicted values: darker shades typically correspond to higher predicted values, while lighter shades correspond to lower predicted values.}
	\label{fig:rplotcars}
\end{figure}

The E2Tree output related to the Auto MPG dataset is presented in Figure~\ref{fig:rplotcars}, provides a graphical representation of the E2Tree structure,   and Table~\ref{tab_res_cars} presents the node-specific numerical results for the E2Tree. The results show the comprehensive nature of our proposal, highlighting the entire decision paths and providing crucial information at each node for predictive analysis.  Table~\ref{tab_res_cars} outlines the primary information and specifies the prediction paths made by the RF. One of the significant advantages of the E2Tree model, as depicted in Figure~\ref{fig:rplotcars}, is the immediate visualization of the importance of splits. This straightforward visualisation explains the relational structure between the response variable and the predictors, along with their interactions, which cannot be directly obtained from a fitted RF model. Figure~\ref{fig:rplotcars} reveals negative effects of displacement, weight and horsepower, and positive effects of model\_year, origin and horsepower on the outcome variable MPG.
We want to stress that one of the most important features of the E2Tree model is its ability to provide clear interpretations of the “If-Then” pathways. These pathways show how interactions between predictors culminate in the final classification at each terminal node. This interpretative capability is provided through both graphical and tabular representations, thus offering a comprehensive understanding of the predictive mechanisms within the model.

\begin{sidewaystable}
    \centering
    \begin{adjustbox}{max width=\textwidth}
    \begin{tabular}{|l|l|l|l|l|l|l|}
    \hline
        $t$ & $n_{t}$ & Prediction & $W_{t_{ij}}$ & $s^*$ & Node Type & Path \\ 
        \hline
        1 & 278 & 23,33 & 0,54 & displacement $\leq$ 267 & Non-Terminal & ~ \\ \hline
        2 & 208 & 26,20 & 0,54 & displacement $\leq$ 183 & Non-Terminal & displacement $\leq$ 267 \\ \hline
        4 & 155 & 28,59 & 0,54 & weight $\leq$ 2074 & Non-Terminal & displacement $\leq$ 267 \& displacement $\leq$ 183 \\ \hline
        8 & 42 & 32,76 & 0,12 & model\_year $\leq$ 77 & Non-Terminal & displacement $\leq$ 267 \& displacement $\leq$ 183 \& weight $\leq$ 2074 \\ \hline
        16 & 24 & 30,33 & 0,36 & origin $\leq$ 2 & Non-Terminal & displacement $\leq$ 267 \& displacement $\leq$ 183 \& weight $\leq$ 2074 \& model\_year $\leq$ 77 \\ \hline
        32 & 14 & 29,07 & 0,00 & horsepower $\leq$ 46 & Non-Terminal & \parbox{13cm}{displacement $\leq$ 267 \& displacement $\leq$ 183 \& weight $\leq$ 2074 \& model\_year $\leq$ 77 \& origin $\leq$ 2} \\ \hline
        64 & 3 & 25,67 & 0,84 & ~ & Terminal & \parbox{13cm}{displacement $\leq$ 267 \& displacement $\leq$ 183 \& weight $\leq$ 2074 \& model\_year $\leq$ 77 \& origin $\leq$ 2 \& horsepower $\leq$ 46} \\ \hline
        65 & 11 & 30,00 & 0,10 & ~ & Terminal & \parbox{13cm}{displacement $\leq$ 267 \& displacement $\leq$ 183 \& weight $\leq$ 2074 \& model\_year $\leq$ 77 \& origin $\leq$ 2 \& horsepower $>$ 46} \\ \hline
        33 & 10 & 32,10 & 0,56 & ~ & Terminal & \parbox{13cm}{displacement $\leq$ 267 \& displacement $\leq$ 183 \& weight $\leq$ 2074 \& model\_year $\leq$ 77 \& origin $>$ 2} \\ \hline
        17 & 18 & 35,99 & 0,00 & ~ & Terminal & displacement $\leq$ 267 \& displacement $\leq$ 183 \& weight $\leq$ 2074 \& model\_year $>$ 77 \\ \hline
        9 & 113 & 27,05 & 0,42 & model\_year $\leq$ 78 & Non-Terminal & displacement $\leq$ 267 \& displacement $\leq$ 183 \& weight $>$  2074 \\ \hline
        18 & 62 & 23,76 & 0,66 & displacement $\leq$ 113 & Non-Terminal & displacement $\leq$ 267 \& displacement $\leq$ 183 \& weight $>$  2074 \& model\_year $\leq$ 78 \\ \hline
        36 & 23 & 25,65 & 0,25 & ~ & Terminal & \parbox{13cm}{displacement $\leq$ 267 \& displacement $\leq$ 183 \& weight $>$  2074 \& model\_year $\leq$ 78 \& displacement $\leq$ 113} \\ \hline
        37 & 39 & 22,64 & 0,59 & ~ & Terminal & \parbox{13cm}{displacement $\leq$ 267 \& displacement $\leq$ 183 \& weight $>$  2074 \& model\_year $\leq$ 78 \& displacement $>$ 113} \\ \hline
        19 & 51 & 31,04 & 0,00 & displacement $\leq$ 108 & Non-Terminal & displacement $\leq$ 267 \& displacement $\leq$ 183 \& weight $>$  2074 \& model\_year $>$  78 \\ \hline
        38 & 15 & 36,01 & 0,00 & ~ & Terminal & \parbox{13cm}{displacement $\leq$ 267 \& displacement $\leq$ 183 \& weight $>$  2074 \& model\_year $>$  78 \& displacement $\leq$ 108} \\ \hline
        39 & 36 & 28,97 & 0,05 & horsepower $\leq$ 105 & Non-Terminal & \parbox{13cm}{displacement $\leq$ 267 \& displacement $\leq$ 183 \& weight $>$  2074 \& model\_year $>$  78 \& displacement $>$ 108} \\ \hline
        78 & 31 & 29,55 & 0,00 & ~ & Terminal & \parbox{13cm}{displacement $\leq$ 267 \& displacement $\leq$ 183 \& weight $>$  2074 \& model\_year $>$  78 \& displacement $>$  108 \& horsepower $\leq$ 105} \\ \hline
        79 & 5 & 25,38 & 0,00 & ~ & Terminal & \parbox{13cm}{displacement $\leq$ 267 \& displacement $\leq$ 183 \& weight $>$  2074 \& model\_year $>$  78 \& displacement $>$  108 \& horsepower $>$ 105} \\ \hline
        5 & 53 & 19,21 & 0,55 & model\_year $\leq$ 75 & Non-Terminal & displacement $\leq$ 267 \& displacement $>$  183 \\ \hline
        10 & 24 & 17,75 & 0,74 & ~ & Terminal & displacement $\leq$ 267 \& displacement $>$  183 \& model\_year $\leq$ 75 \\ \hline
        11 & 29 & 20,41 & 0,00 & ~ & Terminal & displacement $\leq$ 267 \& displacement $>$  183 \& model\_year $>$  75 \\ \hline
        3 & 70 & 14,78 & 0,79 & model\_year $\leq$ 76 & Non-Terminal & displacement $>$  267 \\ \hline
        6 & 54 & 13,87 & 0,87 & weight $\leq$ 3777 & Non-Terminal & displacement $>$  267 \& model\_year $\leq$ 76 \\ \hline
        12 & 13 & 15,38 & 0,45 & displacement $\leq$ 318 & Non-Terminal & displacement $>$  267\& model\_year $\leq$ 76 \& weight $\leq$ 3777 \\ \hline
        24 & 8 & 16,25 & 0,48 & ~ & Terminal & displacement $>$  267 \& model\_year $\leq$ 76 \& weight $\leq$ 3777 \& displacement $\leq$ 318 \\ \hline
        25 & 5 & 14,00 & 0,00 & ~ & Terminal & displacement $>$  267 \& model\_year $\leq$ 76 \& weight $\leq$ 3777 \& displacement $>$  318 \\ \hline
        13 & 41 & 13,39 & 0,88 & ~ & Terminal & displacement $>$  267 \& model\_year $\leq$ 76 \& weight $>$  3777 \\ \hline
        7 & 16 & 17,84 & 0,11 & weight $\leq$ 3880 & Non-Terminal & displacement $>$  267 \& model\_year $>$  76 \\ \hline
        14 & 8 & 19,20 & 0,00 & ~ & Terminal & displacement $>$  267 \& model\_year $>$  76 \& weight $\leq$ 3880 \\ \hline
        15 & 8 & 16,49 & 0,72 & ~ & Terminal & displacement $>$  267 \& model\_year $>$  76 \& weight $>$  3880 \\ \hline
    \end{tabular}
    \end{adjustbox}
    \caption{Auto MPG data: results of E2Tree.}
    \label{tab_res_cars}
\end{sidewaystable}

\begin{figure}[H]
\centering
\begin{subfigure}{0.45\textwidth}
  \centering
  \includegraphics[scale=0.125]{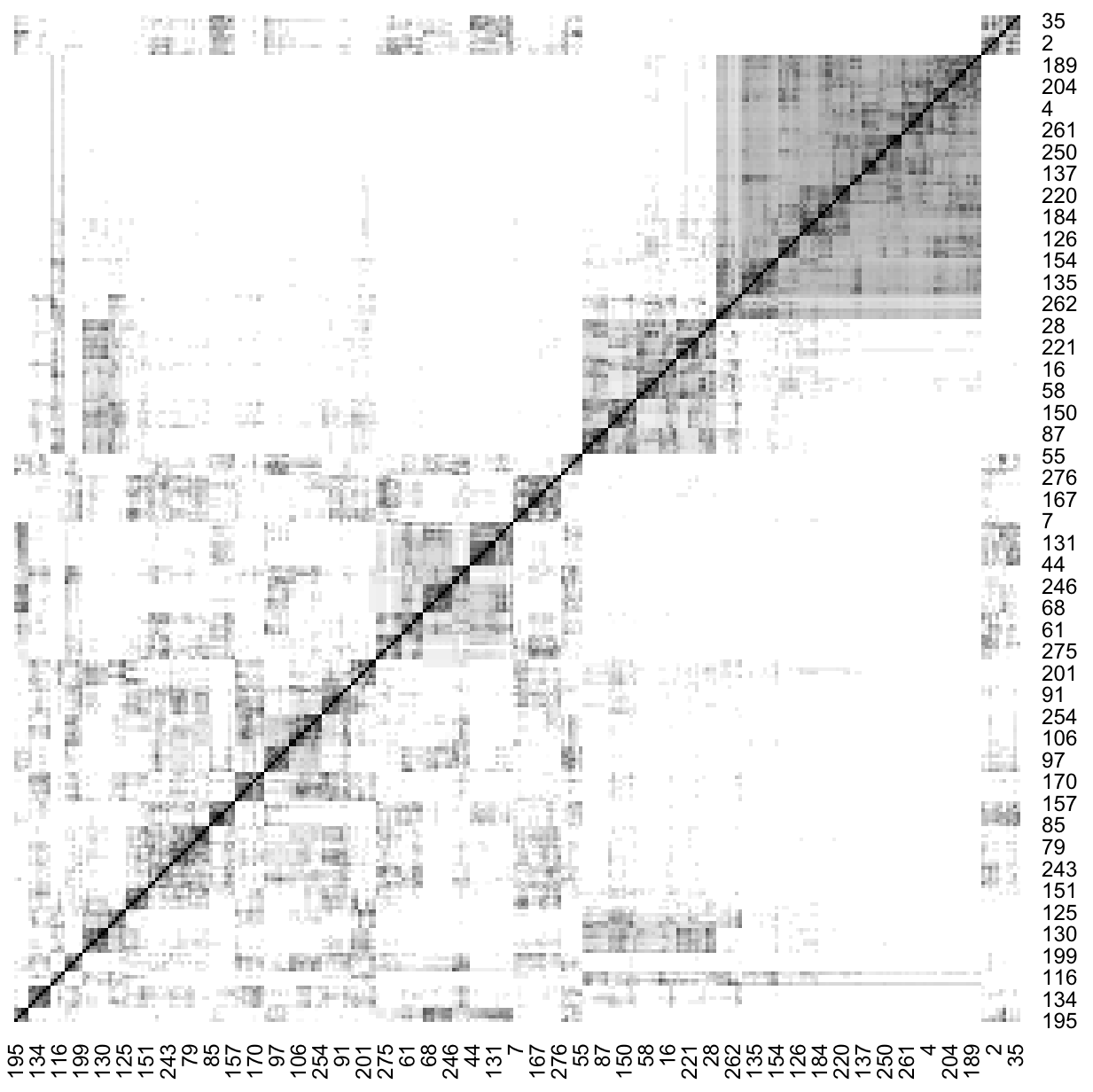}
  \caption{}
\end{subfigure}
\hfill
\begin{subfigure}{0.45\textwidth}
  \centering
  \includegraphics[scale=0.125]{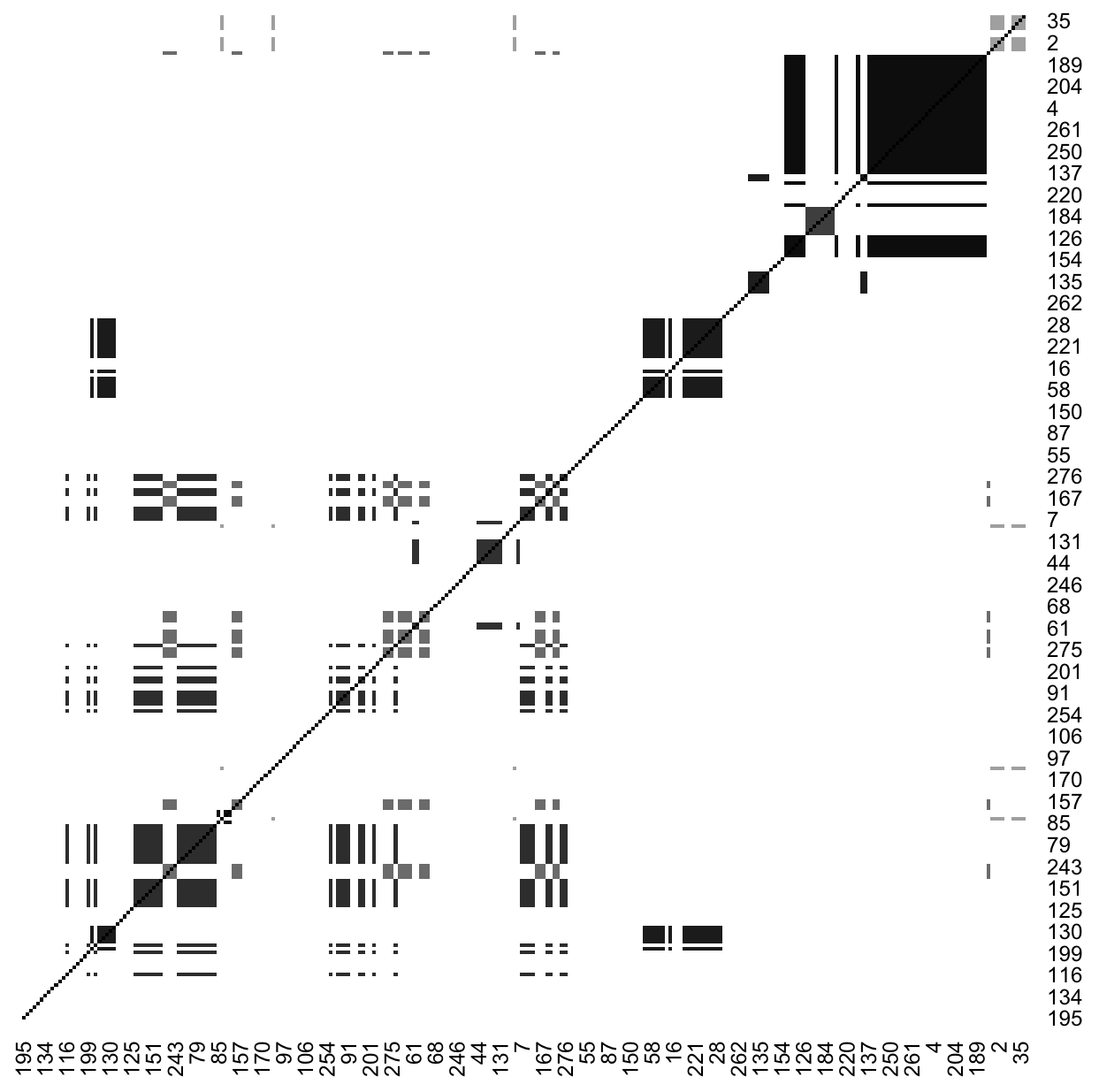}
  \caption{}
\end{subfigure}
\caption{Auto MPG data: heatmap of the matrix $O_{ij}$ (panel a) and heatmap of the matrix $\hat{O}_{ij}$ estimated by E2Tree (panel b).}
\label{comparison2}
\end{figure}

Figure~\ref{comparison2} depicts the two heatmaps corresponding to the matrix $O_{ij}$ derived from the RF output and the matrix estimated by the E2Tree algorithm, which is denoted as $\hat{O}_{ij}$. 
As illustrated in Figure~\ref{comparison2}, the E2tree algorithm is capable of reconstructing the structure of the $O_{ij}$ matrix of the RF, despite the Auto MPG dataset containing more predictors than the Iris dataset used for comparison. Further, it addresses the issue of predicting fuel consumption, which is influenced by a number of external factors. Thirdly, the dataset exhibits non-linear relationships between the variables. The graphical comparison of the two heatmaps demonstrates that the matrix $\hat{O}_{ij}$ is capable of reconstructing the matrix structure obtained from the RF. This is supported by the results of the FMI, which were computed on hierarchical cluster analysis on $O_{ij}$ and $O_{ij}$ with complete-linkage criterion and $k=11$. The FMI is approximately equal to 75\%, which illustrates that E2Tree is able to explain the structure present in the RF model.

%In Figure~\ref{comparison2} the two heatmaps are shown. 
%%we measured how well the E2Tree estimated matrix $\hat{O}_{ij}$ was able to reconstruct the structure of the RF $O_{ij}$ matrix \\
%Since the Auto MPG dataset is more complex than the Iris toy example. Firstly, it contains a larger number of variables and observations. Secondly, it addresses the issue of predicting fuel consumption, which is influenced by multiple external factors. Thirdly, it contains non-linear relationships between the variables. Thus, the graphical comparison of the two matrix heatmaps shows that the matrix $\hat{O}_{ij}$ could reconstruct the matrix structure obtained from the RF. This is evidenced by the Fowlkes-Mallows index, which is approximately 75\%. This demonstrates that E2Tree was able to capture the structure present in the RF model, as there is a high degree of similarity between $O_{ij}$ and $\hat{O}_{ij}$. 

\section{Conclusions}
\label{sec:conclusions}

Proposed for classification tasks, E2Tree offers a coherent, tree-like structure that facilitates an in-depth understanding of the relationships between features and their corresponding predictions while preserving the model’s predictive accuracy. We have now extended the Explainable Ensemble Trees (E2Tree) to the regression context. Our primary aim is to explain and graphically represent the relationships and interactions between the variables used in ensemble methods, such as the RF algorithm. E2Tree now addresses both interpretability and explainability issues in both classification and regression contexts. It provides both local and global explanations, allowing users to understand the decision-making process of the model at a local observation level and a global general level. This dual capability ensures that users can see the detailed logic for specific instances as well as the patterns and rules that the black-box model follows. E2Tree ensures that machine learning models remain tools of empowerment, offering an understandable decision-making process. Our proposal ensures the following advantages: it interprets and explains the ensemble process through a single tree-like structure, maintains the accuracy performance of the RF ensemble, and provides a powerful tool to explain interactions among predictors determining the final node predictions and classifications. Our further aim is to generalize the approach to other ensemble-tree methodologies. Additionally, we are developing metrics to assess the fidelity of our results against the outputs of the ensemble decision tree, ensuring consistency and accuracy in our interpretations.

\backmatter

\bmhead{Acknowledgements}
This research has been financed by the following research projects: PRIN-2022 "SCIK-HEALTH" (Project Code: 2022825Y5E; CUP: E53D2300611006); PRIN-2022 PNRR "The value of scientific production for patient care in Academic Health Science Centres" (Project Code: P2022RF38Y; CUP: E53D23016650001).

\section*{Declarations}

\bmhead{Author contribution}
Massimo Aria, Agostino Gnasso: Conceptualization, Supervision; 
Massimo Aria, Agostino Gnasso, Carmela Iorio: Methodology; 
Agostino Gnasso: Data collection, Data analysis, Writing-Original draft preparation; 
Massimo Aria, Agostino Gnasso, Carmela Iorio, Marjolein Fokkema: Writing-Reviewing and Editing

\bmhead{Conflict of interest}
No potential conflict of interest was reported by the author(s).

\bmhead{Data availability}
The data are freely available on the platform from where it was extracted, and the dataset in specific can be requested from the corresponding author with a justification for the need to access the dataset.

\bmhead{Code availability}
The code used for the analyses presented in this work is available upon request from the corresponding author. All analyses were performed using the \textit{e2tree} package, which is available on the GitHub repository \url{https://github.com/massimoaria/e2Tree}.

%\begin{itemize}
%\item Funding
%\item Conflict of interest 
%\item Ethics approval and consent to participate
%\item Consent for publication
%\item Data availability 
%\item Materials availability
%\item Code availability 
%\item Author contribution
%\end{itemize}

%\bibliographystyle{apalike}
\bibliography{biblio_reg} % common bib file
%% if required, the content of .bbl file can be included here once bbl is generated
%\input sn-article.bbl

\end{document}